\documentclass[preprints,article,accept,moreauthors,pdftex,10pt,a4paper]{Definitions/mdpi}

\firstpage{1} 
\pubvolume{10}
\issuenum{1}
\articlenumber{82}
\pubyear{2019}
\copyrightyear{2019}
\history{Received: date; Accepted: date; Published: date}

\pdfoutput=1

\usepackage{natbib}
\usepackage{amsmath,amsfonts,amssymb,amsthm}
\usepackage{siunitx}
\usepackage{bbm}
\usepackage{microtype}
\usepackage{cleveref}
\usepackage{todonotes}

\DeclareMathOperator*{\argmax}{arg\,max}

\newcommand{\numTwo}[1] {\num[round-precision=2]{#1}}

\Title{Machine Reading Comprehension for Answer Re-Ranking in Customer Support Chatbots}

\Author{Momchil Hardalov $^{1,*}$, Ivan Koychev $^{1}$ and Preslav Nakov $^{2}$}
\AuthorNames{Momchil Hardalov, Ivan Koychev and Preslav Nakov}

\address{%
$^{1}$ \quad Faculty of Mathematics and Informatics, Sofia University, 1164 Sofia, Bulgaria; koychev@fmi.uni-sofia.bg\\
$^{2}$ \quad Qatar Computing Research Institute, Hamad Bin Khalifa University, 34110 Doha, Qatar; pnakov@qf.org.qa}

\corres{Correspondence: hardalov@fmi.uni-sofia.bg}

\abstract{Recent advances in deep neural networks, language modeling and language generation have introduced new ideas to the field of conversational agents. As a result, deep neural models such as sequence-to-sequence, memory networks, and the Transformer have become key ingredients of state-of-the-art dialog systems. While those models are able to generate meaningful responses even in unseen situations, they need a lot of training data to build a reliable model.  Thus, most real-world systems have used traditional approaches based on information retrieval (IR) and even hand-crafted rules, due to their robustness and effectiveness, especially for narrow-focused conversations. Here, we present a method that adapts a deep neural architecture from the domain of machine reading comprehension to re-rank the suggested answers from different models using the question as a context. We train our model using negative sampling based on question--answer pairs from the Twitter Customer Support Dataset.
The experimental results show that our re-ranking framework can improve the performance in terms of word overlap and semantics both for individual models as well as for model combinations.}


\begin{document}

\section{Introduction}

The growing popularity of smart devices, personal assistants, and online customer support systems has driven the research community to develop various new methodologies for automatic question answering and chatbots. In the domain of conversational agents, two general types of systems have become dominant: ({i})~retrieval-based, and ({ii})~generative. While the former produce clear and smooth output, the latter bring flexibility and the ability to generate new unseen answers. 

In this work, we focus on finding the most suitable answer for a question, where each candidate can be produced by a different system, e.g., knowledge-based, rule-based, deep neural network, retrieval, etc.
In particular, we propose a re-ranking framework based on machine reading comprehension \cite{yu2018qanet, seo2016bi-directional, Chen:2017:DrQA} for question--answer pairs. Moreover, instead of selecting the top candidate from the re-ranker's output, we use probabilistic sampling that aims to diversify the agent’s language and to up-vote popular answers from different input models. We train our model using negative sampling based on question--answer pairs from the Twitter Customer Support Dataset.

In our experimental setup, we adopt a real-world application scenario, where we train on historical logs for some period of time, and then we test on logs for subsequent days.
We evaluate the model using both semantic similarity measures, as well as word-overlap ones such as BLEU \cite{papineni2002bleu} and ROUGE~\cite{lin2004rouge}, which come from machine translation and text summarization.

\newpage

The remainder of this paper is organized as follows: 
Section~\ref{sec:related} presents some related work in the domain of conversational agents and answer combination.
Section~\ref{sec:model} describes our framework and the general workflow for answer re-ranking.
Section~\ref{sec:data} introduces the original dataset and explains how we used in to build a new, task-specific one with negative sampling; it also offers insights about the dialogs and the pre-processing.
Section~\ref{sec:experiments} describes our experiments, and gives details about the training parameters. 
Section~\ref{sec:resanddis} presents the performance of each model and discusses the results. 
Finally, Section~\ref{sec:conclus} concludes and suggests possible directions for future work.

\section{Related Work}
\label{sec:related}

\subsection{Conversational Agents}

The emergence of large conversational corpora such as the Ubuntu Dialog corpus~\citep{lowe2015ubuntu}, OpenSubtitles~\citep{LISON16.947:opensubs}, CoQA~\citep{reddy2018coqa} and the Microsoft Research Social Media Conversation Corpus~\citep{Microsoft}  
~has enabled the use of generative models and end-to-end neural networks in the domain of conversational agents. In particular, sequence-to-sequence (seq2seq) models, which were initially proposed for machine translation \citep{luong2015encoder, sutskever2014sequence, bahdanau2014neural}, got adapted to become a standard tool for training end-to-end dialogue systems. Early vanilla seq2seq models \citep{DBLP:journals/corr/VinyalsL15} got quickly extended to model hierarchical structure \citep{serban2015hierarchical}, context~\citep{conf/naacl/SordoniGABJMNGD15}, and combination thereof  \citep{Sordoni:2015:HRE:2806416.2806493}. While models were typically trained on corpora such as Ubuntu, some work \citep{boyanov-EtAl:2017:RANLP} has also used data from Community Question Answering forums \citep{nakov2017semeval}; this means forming a training pair involving a question and each good answer in the corresponding question-answer thread.

More recently, the Transformer, a model without recurrent connections, was proposed \citep{NIPS2017_7181:transformer}, demonstrating state-of-the-art results for Machine Translation in various experimental scenarios for several language pairs and translation directions, thus, emerging as a strong alternative to seq2seq methods. The fact that it only uses self-attention makes it a lot faster both at training and at inference time, even though its deep architecture requires more calculations than a seq2seq model, it enables high degree of parallelism, while maintaining the ability to model word sequences through the mechanism of attention and positional embeddings.

In the domain of customer support, it has been shown that generative models such as seq2seq and the Transformer perform better then retrieval-based models, but they fail in the case of insufficient training data \citep{hardalov:10.1007/978-3-319-99344-7_5:customer}. Other works have incorporated intent categories and semantic matching into an answer selection model, which uses a knowledge base as its source \citep{LI:10.1007/978-3-319-99495-6_1:customerqa}. 
In the insurance domain,~\citet{Feng:2015:ASRU}~proposed a generic deep learning approach for answer selection, based on convolutional neural networks (CNN) \cite{lecunn:1989:neural}. In \citet{li2015recurrent} combined recurrent neural networks based on long short-term memory (LSTM) cells \cite{Hochreiter:1997:LSM:1246443.1246450} and reinforcement learning (RL) to learn without the need of prior domain knowledge.

\subsection{Answer Combination}

Answer combination has been recognized as an important research direction in the domain of customer support chatbots. For example,
\citet{qiu2017alime} used an attentive seq2seq re-ranker to choose dynamically between the outputs of a retrieval-based and a seq2seq model. Similarly, \citet{cui2017superagent} combined a fact database, FAQs, opinion-oriented answers, and a neural-based chit-chat generator, by training a meta-engine that chooses between them. 

Answer combination is also a key research topic in the related field of information retrieval (IR). For example, \citet{Pang:2017:DND:3132847.3132914} proposed a generic relevance ranker based on deep learning and CNNs~\cite{lecunn:1989:neural}, which tries to maintain standard IR search engine characteristics, such as exact matching and query term importance, while enriching the results based on semantics, proximity heuristics, and~diversification.

\section{Re-Ranking Model}
\label{sec:model}

\begin{figure}[h]
    \centering
    \includegraphics[scale=0.38]{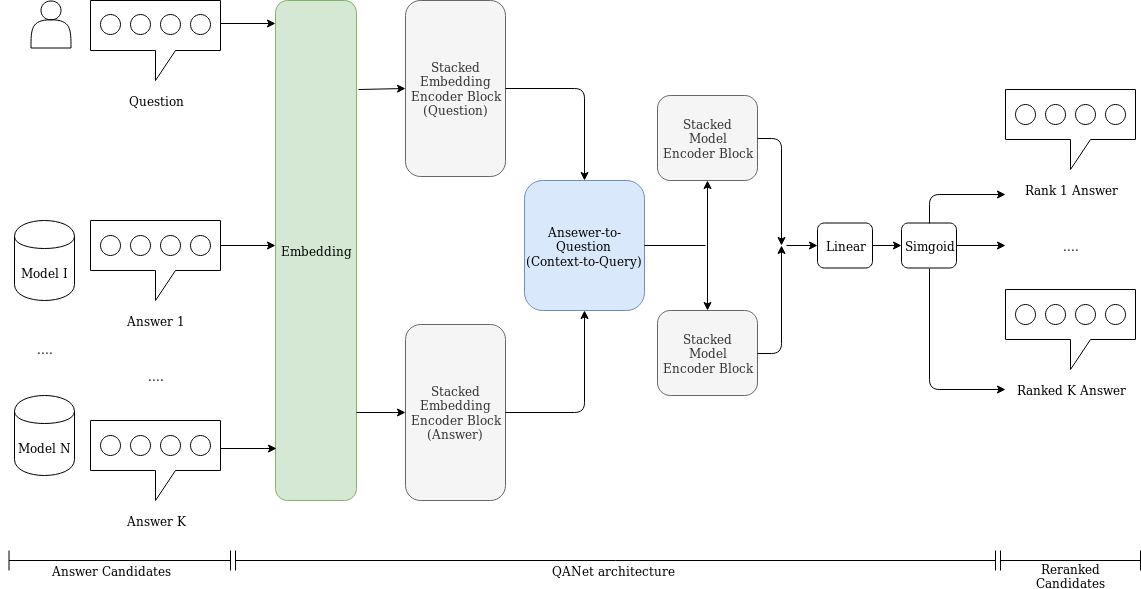}
    \caption{Our answer re-ranking framework, based on the QANet architecture.}
    \label{fig:qanet}
\end{figure}

Our re-ranking framework uses a classifier based on QANet \cite{yu2018qanet}, a state-of-the-art architecture for machine reading comprehension, to evaluate whether a given answer is a good fit for the target question. 
It then uses the posterior probabilities of the classifier to re-rank the candidate answers, as shown in Figure~\ref{fig:qanet}. 

\subsection{Negative Sampling}

Our goal is to distinguish ``good'' vs. ``bad'' answers, but the original dataset only contains valid, i.e., ``good'' question--answer pairs. 
Thus, we use \emph{negative sampling}
  \cite{NIPS2013:w2v}, where we replace the original answer to the target question with a random answer from the training dataset.
We further compare the word-based cosine similarity between the original and the sampled answer, and, in some rare cases, we turn a ``bad'' answer into ``good'' one if it is too similar to the original ``good'' answer.

\subsection{QANet Architecture}
\label{sec:qanet}
Machine reading comprehension aims to answer a question by looking to extract a string from a given text context. Here, we use that model to measure the goodness of a given question--answer pair.

The first layer of the network is a standard an embedding layer, which transforms words into low-dimensional dense vectors.  Afterwards, a two-layer highway network \citep{srivastava2015highway}  is added on top of the embedding representations. This allows the network to regulate the information flow using a gated mechanism. The output of this layer is of dimensionality $\#words \times d$, where $\#words$ is the number of words in the encoded sentence (Note that it differs for the question vs. the answer. See Section \ref{subsec:preprocess} for more detail.) and $d$ is the input/output dimensionality of the model for all Transformer layers, which is required by the architecture.

We experiment with two types of input embeddings. First, we use 200-dimensional GloVe \citep{pennington2014glove} vectors trained on 27 billion Twitter posts. We compare their performance to ELMo~\cite{Peters:2018:ELMo}, a recently proposed way to train contextualized word representations. In ELMo, these word vectors are learned activation functions of the internal states of a deep bi-directional language model. The latter is built upon a single (embedding) layer, followed by two LSTM \citep{Hochreiter:1997:LSM:1246443.1246450} layers, which are fed the words from a target sentence in a forward and a backward direction, accordingly. We obtain the final embedding by taking a weighted average over all three layers as suggested in \citep{Peters:2018:ELMo}.

The embedding encoder layer is based on a convolution, followed by self-attention \cite{NIPS2017_7181:transformer} and a~feed-forward network. We use a kernel size of seven, $d$ filters, and four convolutional layers within a block. The output of the layer is $f(layernorm(x)) + x$, where $layernorm$ is the layer normalization operation~\citep{ba2016layer}. The output again is mapped to $\#words \times d$ by a 1D convolution. The input and the embedding layers are learned separately for the question and the answer.

The attention layer is a standard module for machine reading comprehension models. We call it \emph{answer-to-question} (\emph{A2Q}) and \emph{question-to-answer} (\emph{Q2A}) attention, which are also known as \emph{context-query} and \emph{query-context}, respectively. Let us denote the output of the encoder for the question as $Q$ and for the answer as $A$. In order to obtain the attention, the model first computes a matrix $S$ with similarities between each two words for the question and the answer, then the values are normalized using softmax. The similarity function is defined as follows: $f(a,q) = W_0[a; q; a \odot q]$. 

We adopt the notation $\overline{S} = softmax(S)$, which is a softmax normalization over the rows of $S$, and $\overline{\overline{S}} = softmax(S^{\intercal})$ is a normalization over the columns. Then, the two attention matrices are computed as $A2Q = \overline{S} \cdot Q^{\intercal}$, and $Q2A = \overline{S} \cdot \overline{\overline{S}}^{\intercal} \cdot C^{\intercal}$.

The attention layer is followed by a model layer, which takes as input the concatenation of $[a;a2q;a \odot a2q; a \odot q2a]$, where we use small letters to denote rows from the original matrices. For the output layer, we learn two different representations by passing the output of the model layer to two residual blocks, applying dropout~\citep{srivastava2014dropout} only to the inputs of the first one. We predict the output as  $P(a|q) = \sigma(W_o[M_0;M_1])$. The weights are learned by minimizing a binary cross-entropy loss.

\subsection{Answer Selection}

We experimented with two answer selection strategies: ({i})~max, and ({ii})~proportional sampling after softmax normalization. The former strategy is standard and it selects the answer with the highest score, while the latter one returns a random answer with probability proportional to the score returned by the softmax, aiming at increasing the variability of the answers.

For both strategies, we use a linear projection applied on the output of the last residual model block, which is shows as ``linear block'' in Figure~\ref{fig:qanet}. We can generalize the latter as follows: $o(q, a_k) = W_o[M]$, where $M$ is the concatenation of the outputs of one or more residual model blocks. 

We present the formulation of the two strategies, as we introduce the following notation: $Ans$ is the selected utterance by the agent; $o(q, a_k)$ is the output of the model before applying the sigmoid function; $q$ is the original question by the user; $A$ is the set of possible answers that we want to re-rank. Equation~(\ref{eq:greedyans}) shows the selection process in the max case.

\begin{equation}
Ans = \argmax_{a \in A}(o(q, a))
\label{eq:greedyans}
\end{equation}

We empirically found  that the answer selection based on the \emph{max} strategy does not always perform well. As our experimental results in Tables \ref{tab:modelressingle} and \ref{tab:modelrescomb} show, we can gain notable improvement by using proportional sampling after softmax normalization, instead of always selecting the answer with the highest probability. In our experiments, we model $Ans$ as a random variable that follows a categorical distribution over $K = |A|$ events (candidate answers). For each of the question--answer pairs ($q$, $a$), we define the probability $p$ that $a$ is a good answer to $q$ using \textup{softmax} as shown in Equations (\ref{eq:softmaxnorm}) and (\ref{eq:catans}). Finally, we draw a random sample from Equation (\ref{eq:catans}) to obtain the best matching answer.

\begin{equation}
p|q,A \sim \textup{softmax}({o(q, a_1), \cdots, o(q, a_K)})
\label{eq:softmaxnorm}
\end{equation}

\begin{equation}
Ans|p \sim \textup{Cat}(K, p)
\label{eq:catans}
\end{equation}

\section{Data}
\label{sec:data}

The data and the resources that could be used to train customer support conversational agents are generally very scarce, as companies keep conversations locked on their own proprietary support systems. This is due to customer privacy concerns and to companies not wanting to make public their know-how and the common issues about their products and services. An extensive 2015 survey on available dialog corpora by \citet{serban2018survey} found no good publicly available dataset for real-world customer support.

In early 2018, this situation changed as a new open dataset for Customer Support on Twitter~\citep{Twitter} was made available on Kaggle.
~It contains 3M tweets and replies 
for twenty big companies such as Amazon, Apple, Uber, Delta, and Spotify, among others. 
As customer support topics from different organizations are generally unrelated to each other, we focus only on tweets related to Apple support, which represents the largest number of tweets in the~corpus. 

We filtered all utterances that redirect the user to another communication channel, e.g.,~direct messages, which are not informative for the model and only bring noise. Moreover, since answers evolve over time, we divided our dataset into a training and a testing part, keeping earlier posts for training and the latest ones for testing. We further excluded from the training set all conversations that are older then sixty days. For evaluation, we used dialogs from the last five days in the dataset, to simulate a real-world scenario for customer support. We ended up with a dataset of 49,626 question--answer pairs divided into 45,582 for training and 4,044 for testing. Finally, we open-sourced our code for pre-processing and filtering the data, making it available to the research community \citep{chatbot}.

Table~\ref{tab:dialogstats} shows some statistics about our dataset. On the top of the table, we can see that the average number of turns per dialog is under three, which means that most of the dialogues finish after one answer from the customer support. The bottom of the table shows the distribution of the words in the user questions vs. the customer support answers. We can see that answers tend to be slightly longer, which is natural as replies by customer support must be extensive and helpful.

\begin{table}[H]

  \centering
    \caption[Statistics about our dataset.]{Statistics about our dataset. (Reprinted by permission from Springer Nature: Springer Lecture Notes in Computer Science (Hardalov,  M.;  Koychev,  I.;  Nakov,  P.
Towards  Automated  Customer  Support \cite{hardalov:10.1007/978-3-319-99344-7_5:customer}), 2018)}

\begin{tabular}{>{\arraybackslash}p{3cm} >{\centering\arraybackslash}p{2cm} >{\centering\arraybackslash}p{2cm}}
	\toprule
	& \bf Questions & \bf Answers \\
    \midrule
    Avg. \# words      & $\numTwo{21.314814}$ &    $\numTwo{25.884436}$ \\
    Min \# words      &  $\numTwo{1.000000}$ &    $\numTwo{3.000000}$ \\
    1st quantile (\#words)  &   $\numTwo{13.000000}$ &    $\numTwo{20.000000}$ \\ 
    Mode (\# words) &     $\numTwo{20.000000}$ &   $\numTwo{23.000000}$ \\
    3rd quantile (\#words) &    $\numTwo{ 27.000000}$ &   $\numTwo{29.000000}$ \\
    Max \# words   &     $\numTwo{136.000000}$ &  $\numTwo{70.000000}$ \\
    \midrule
%
  \multicolumn{3}{c}{\bf Overall}\\
    \midrule
    \multicolumn{2}{c}{\# question--answer pairs} & 49,626 \\
    \multicolumn{2}{c}{\# words (in total)} &  26,140 \\
    \multicolumn{2}{c}{Min \# turns per dialog} &  $\num[round-mode = places]{2}$ \\
    \multicolumn{2}{c}{Max \# turns per dialog} &  $\num[round-mode = places]{106}$ \\
    \multicolumn{2}{c}{Avg. \# turns per dialog} & 2.6 \\
	\midrule
    \multicolumn{2}{c}{Training set: \# of dialogs} & 45,582 \\
    \multicolumn{2}{c}{Testing set: \# of dialogs} & 4,044 \\
    \bottomrule
    \\
\end{tabular}

  \label{tab:dialogstats}
\end{table}

\section{Experiments}
\label{sec:experiments}

\subsection{Preprocessing}
\label{subsec:preprocess}

Since Twitter has its own specifics of writing in terms of both length (by design, tweets have been strictly limited to 140 characters; this constraint has been relaxed to 280 characters in 2017) and style, standard text tokenization is generally not suitable for tweets. Therefore, we used a specialized Twitter tokenizer \cite{manning2014stanford} to preprocess the data. Then, we further replaced shorthand entries such as \emph{'ll}, \emph{'d}, \emph{'re}, \emph{'ve}, with the most corresponding literary form, e.g., \emph{will}, \emph{would}, \emph{are}, \emph{have}. We also replaced shortened slang words, e.g., \emph{'bout} and \emph{'til}, with the standard words, e.g., \emph{about} and \emph{until}. Similarly, we replaced URLs with the special word \textit{<url>}, all user mentions with \textit{<user>}, and all hashtags with \textit{<hashtag>}. 

Due to the nature of writing in Twitter and the free form of the conversation, some of the utterances contain emoticons and emojis. They are handled automatically by the Twitter tokenizer and treated as a single token. We keep them in their original form, as they can be very useful for detecting emotions and sarcasm, which pose serious challenges for natural language understanding.

Based on the statistics presented in Section~\ref{sec:data}, we chose to trim the length of the questions and of the answers to 60 and 70 words, respectively.

\subsection{Training Setup}

For training, we use the Adam \cite{kingma2015adam} optimizer with decaying learning rate, as implemented in TensorFlow \cite{abadi2016tensorflow}. We start with the following values: learning rate $\eta = \num{5e-04}$, exponential decay rate for the 1st and the 2nd momentum $\beta_1 = \num[round-precision=1]{0.9}$ and $\beta_2 = \num{0.999}$, and constant for prevention of division by zero $\epsilon = \num{1e-7}$.
Then, we decay the learning after each epoch by a factor of 0.99. We also apply dropout with a probability of 0.1, and L2 weight decay on all trainable variables with $\lambda = \num{3e-7}$. We train each model for 42K steps with a batch size of 64. We found these values by running a grid search on a dev set (extracted as a fraction of the training data) and using the values suggested in \citep{yu2018qanet}, where applicable.

\subsection{Individual Models}
\label{subsec:singlemodels}

Following \citep{hardalov:10.1007/978-3-319-99344-7_5:customer}, we experiment with three individual models: ({i})~Information Retrieval-based (IR), ({ii})~Sequence-to-sequence (seq2seq) and ({iii})~the Transformer. 

For {IR}, we use ElasticSearch \citep{ElasticSearch} with  English analyzer enabled, whitespace- and punctuation-based tokenization, and word 3-grams. We further use the default BM25 algorithm~\cite{Robertson:2009:PRF:1704809.1704810}, which is an improved version of TF.IDF. For all training questions and for all testing queries, we~append the previous turns in the dialog as context.

For {seq2seq}, we use a bi-directional LSTM network with 512 hidden units per direction. The~decoder has two uni-directional layers connected directly to the bi-directional layer in the encoder. The network takes as input words encoded as 200-dimensional embeddings. It is a combination of pre-trained GloVe \cite{pennington2014glove} vectors for the known words, and a positional embedding layer, learned as model parameters, for the unknown words. The embedding layers for the encoder and for the decoder are not shared, and are learned separately. This separation is due to the words used in utterances by the customers being very different from the posts by the customer support. 

For the {Transformer}, we use two identical layers for the encoder and for the decoder, with four heads for the self-attention. The dimensionality of the input and of the output is $d_{model} = 256$, and the inner dimensionality is $d_{inner} = 512$. The input consists of queries with keys of dimensionality $d_k = 64$ and values of the same dimensionality $d_v = 64$. The input and the output embedding are learned separately with sinusoidal positional encoding.

\subsection{Evaluation Measures}

How to evaluate a chatbot is an open research question. As the problem is related to machine translation (MT) and text summarization (TS), which are nowadays also addressed using seq2seq models, researchers have been using MT and TS evaluation measures such as BLEU~\cite{papineni2002bleu} and ROUGE~\cite{lin2004rouge}, which focus primarily on word overlap and measure the similarity between the chatbot's response and the gold answer to the user question (here, the answer by the customer support).
However, it has been argued \cite{liu-EtAl:2016:EMNLP20163,lowe-EtAl:2017:Long} that such word-overlap measures are not very suitable for evaluating chatbots. Thus, we adopt three additional measures, which are more semantic in nature.

The \textit{embedding average} \citep{lowe2015ubuntu} constructs a vector for a piece of text by taking the average of the word embeddings of its constituent words. Then, the vectors for the chatbot response and for the gold human answer are compared using the cosine similarity.

The \textit{greedy matching} was introduced in the context of intelligent tutoring systems \cite{rus2012comparison}. It matches each word in the chatbot's output to the most similar word in the gold human response, where the similarity is measured as the cosine between the corresponding word embeddings, multiplied by a weighting term (which we set to 1), as shown in Equation (\ref{eq:matching}). Since this measure is asymmetric, we also calculate it with the arguments swapped, and then we take the average as shown in Equation~(\ref{eq:matching_simetric}).

\begin{equation}
\label{eq:matching}
greedy(u_1, u_2) = \frac{
	\sum_{v \in u_1}{} weight(v) * \max_{w \in u_2} cos(v,w)}
    {
    \sum_{v \in u_1}{} weight(v)}
\end{equation}

\begin{equation}
\label{eq:matching_simetric}
simGreedy(u_1, u_2) = \frac{greedy(u_1, u_2) + greedy(u_2, u_1)}{2}
\end{equation}

The \emph{vector extrema} \citep{forgues2014bootstrapping} was proposed for dialogue systems. Instead of averaging the word embeddings of the words in a piece of text, it takes the coordinate-wise maximum (or minimum), as shown in Equation (\ref{eq:exterma}). Finally, the resulting vectors for the chatbot output and for the gold human answer are compared using the cosine similarity.

\begin{equation}
\label{eq:exterma}
extrema(u_i) =  
\begin{cases}
    \max u_i, & if \max u_i \geq |\min u_i| \\
    \min u_i, & \text{otherwise}
\end{cases}
\end{equation}

\section{Evaluation Results}
\label{sec:resanddis}

Below, we first discuss our auxiliary classification task, where the objective is to predict which question--answer pair is ``good'', and then we move to the main task of answer re-ranking. 

\subsection{Auxiliary Task: Question--Answer Goodness Classification}

Table~\ref{tab:modelacc} shows the results for the auxiliary task of question--answer goodness classification. The first column is the name of the model. It is followed by three columns showing the type of embedding used, the size of the hidden layer, and the number of heads (see Section \ref{sec:qanet}). The last column reports the accuracy. Since our dataset is balanced (we generate about 50\% positive, and about 50\% negative examples), accuracy is a suitable evaluation measure for this task.
The top row of the table shows the performance for a majority class baseline. 
The following lines show the results for our full QANet-based model when using different kinds of embeddings. 
We can see that contextualized sentence-level embeddings are preferable to using simple word embeddings as in GloVe or token-level ELMo embeddings. 
Moreover, while token-level ELMo outperforms GloVe when the size of the network is small, there is no much difference when the number of parameters grows ($d_{model} = 128$, $\#Heads = 8$).

\begin{table}[H]

    \centering
        \caption{{Auxiliary task:} question--answer goodness classification results.}
    \label{tab:modelacc}
    \begin{tabular}{ccccc}
         \toprule
         \bf Model & \bf Embedding Type & \bf d\_model & \bf Heads & \bf Accuracy \\
         \midrule
         Majority class & -- & -- & -- & $\num{50.515625}$ \\
         \midrule
\multirow{3}{*}{QANet} & \multirow{3}{*}{GloVe} & 64 & 4 & $\num{80.578125}$ \\
         & & 64 & 8 & $\num{82.82812}$ \\
         & & 128 & 8 & $\num{83.421874}$ \\
         \midrule
        \multirow{3}{*}{QANet} &         \multirow{3}{*} {ELMo (token level)} & 64 & 4 &  $\num{82.921875}$ \\
         & & 64 & 8 & $\num{83.875}$ \\
         & & 128 & 8 & $\num{83.484375}$  \\
         \midrule
                \multirow{2}{*} {QANet} &         \multirow{2}{*}{ELMo (sentence level)} & 64 & 8 & $\num{84.09375}$ \\
         & & 128 & 8 & \bf{$\num{85.45312}$}
          \\
         \bottomrule
    \end{tabular}

\end{table}

\subsection{Answer Selection/Generation: Individual Models}

Table~\ref{tab:modelressingle} reports the performance of the individual models: information retrieval (IR), Sequence-to-sequence (seq2seq), and the Transformer (see Section~\ref{subsec:singlemodels} for more details about these models). In our earlier
~work~\citep{hardalov:10.1007/978-3-319-99344-7_5:customer}, we performed these experiments using exactly the same experimental setup. The table is organized as follows: The first column contains the name of the model used to obtain the best answer. The second and the third columns report the word overlap measures: ({i})~BLEU@2, which uses uni-gram and bi-gram matches between the hypothesis and the reference sentence, and ({ii})~ROUGE-L \cite{lin-och:2004:ACL}, which uses Longest Common Subsequence (LCS). The last three columns are for the semantic similarity measures: ({i})~Embedding Average (Emb Avg) with cosine similarity, ({ii})~Greedy Matching (Greedy Match), and ({iii})~Vector Extrema (Vec Extr) with cosine similarity. In the three latter measures, we used the standard pre-trained word2vec embeddings because they are not learned during training, which helps avoid bias, as has been suggested in \cite{liu-EtAl:2016:EMNLP20163,lowe-EtAl:2017:Long}.

We can see in Table~\ref{tab:modelressingle} that the seq2seq model outperforms IR by a margin on all five evaluation measures, which is consistent with previous results in the literature. What is surprising, however, is the relatively poor performance for the Transformer, which trails behind the seq2seq model on all evaluation measures. We hypothesize that this is due to the Transformer having to learn more parameters as it operates with higher-dimensional word embeddings. Overall, the Transformer is arguably slightly better than the IR model, outperforming it on three of the five evaluation measures.

The last row of Table~\ref{tab:modelressingle} is not an individual model; it is our re-ranker applied to the top answers returned by the IR model. In particular, we use \emph{QANet with Sentence level ELMo}
~($d_{model} = 128$, $\#Heads = 8$). We took the top-5 answer candidates (the value of 5 was found using cross-validation on the training dataset) from the IR model, and we selected the best answer based on our re-ranker's scores. We can see that re-ranking yields improvements for all evaluation measures: $+1.18$ on BLEU@2, $+0.93$ on ROUGE\_L, $+1.12$ on Embedding Average, $+0.67$ on Greedy Matching, and +1.64 in Vector Extrema. These results show that we can get sizable performance gains when re-ranking the top-$K$ predictions of a single model; below we will combine multiple models.

\begin{table}[H]    
    \centering
        \caption{Main task: performance of the individual models.}
    \begin{tabular}{cccccc}
        \toprule
        & \multicolumn{2}{c}{\bf Word Overlap} & \multicolumn{3}{c}{\bf Semantic Similarity} \\
        \cmidrule(lr){2-3} 
        \cmidrule(lr){4-6} 
        \bf Model &  \bf BLEU@2 & \bf ROUGE\_L  & \bf Emb Avg & \bf Greedy Match & \bf Vec Extr \\
        \midrule
        Transformer \cite{hardalov:10.1007/978-3-319-99344-7_5:customer} &  
        $\num[round-mode = places]{12.427277110577412}$ & 
        $\num[round-mode = places]{25.32858959229684}$ &
        $\num[round-mode = places]{75.35035772464175}$ & 
        $\num[round-mode = places]{30.07707395198416}$ &
        $\num[round-mode = places]{39.39895553309098}$ \\
        IR-BM25 \cite{hardalov:10.1007/978-3-319-99344-7_5:customer} &  
        $\num[round-mode = places]{13.732301176466532}$ & 
        $\num[round-mode = places]{22.347860605005877}$ &
        $\num[round-mode = places]{76.52658639738192}$ & 
        $\num[round-mode = places]{29.719912622416917}$ &
        $\num[round-mode = places]{37.98667099511575}$ \\
        seq2seq \cite{hardalov:10.1007/978-3-319-99344-7_5:customer} &  
        $\num[round-mode = places]{15.104511595780986}$ & 
        $\num[round-mode = places]{26.597981802417063}$ &
        $\num[round-mode = places]{77.10925516393053}$ &
        $\num[round-mode = places]{30.809532318667042}$ &
        $\num[round-mode = places]{40.231592168970955}$ \\
        \midrule
        \textbf{QANet on IR (Individual)} &
        $\num{14.918819893} \pm \numTwo{0.128209477}$  & 
        $\num{23.299038571} \pm \numTwo{0.121935742}$  & 
        $\num{77.470088663} \pm \numTwo{0.059103246}$  & 
        $\num{30.396535191} \pm \numTwo{0.059016826}$  & 
        $\num{39.627030115} \pm \numTwo{0.056555891}$  \\         
        \bottomrule
    \end{tabular}

    \label{tab:modelressingle}
\end{table}

\subsection{Main Task: Multi-Source Answer Re-Ranking}

Next, we combine the top-$K$ answers from different models:
IR and seq2seq. We did not include the Transformer in the mix as its output is generative and similar to that of the seq2seq model; moreover, as we have seen in Table~\ref{tab:modelressingle} above, it performs worse than seq2seq on our dataset.  
We set $K = 2$ for the baseline, \emph{Random Top Answer}, 
which selects a random answer from the union of the top $K$ answers by the models involved in the re-ranking. For the remaining re-ranking experiments, we use $K = 5$. We found these values using cross-validation on the training dataset, trying 1--5.

The results are shown in Table~\ref{tab:modelrescomb}, where different representations are separated by a horizontal line. The first row of each group contains the name of the model. Then, on the even rows (second, forth, etc.), we show the results from a greedy answer selection strategy, while on the odd rows are the results from an exploration strategy (softmax sampling). Since softmax sampling and random selection are stochastic in nature, we include a $95\%$ confidence interval for them.

\begin{table}[H]    
    \centering
        \caption{Main task: re-ranking the top $K = 5$ answers returned by the IR and the seq2seq models.}
    \begin{tabular}{cccccc}
        \toprule
       \multirow{2}{*}{\bf Model}  & \multicolumn{2}{c}{\bf Word Overlap} & \multicolumn{3}{c}{\bf Semantic Similarity} \\
        \cmidrule(lr){2-3} 
        \cmidrule(lr){4-6} 
&  \bf BLEU@2 & \bf ROUGE\_L  & \bf Emb Avg & \bf Greedy Match & \bf Vec Extr \\
        \midrule
        \bf Random Top Answer  &  
        $\num{14.516328934} \pm \numTwo{0.123082790}$  & 
        $\num{23.413930960} \pm \numTwo{0.118796211}$  & 
        $\num{77.214657500} \pm \numTwo{0.064671076}$  & 
        $\num{30.244576358} \pm \numTwo{0.065005588}$  & 
        $\num{38.247697434} \pm \numTwo{0.199359940}$  \\ 
        \midrule
        \bf QANet+GloVe & \multicolumn{5}{l}{}    \\
        d=64, h=4 &  
        $\num[round-mode = places]{15.178345469681423}$ & 
        $\num[round-mode = places]{24.127489239551984}$ &
        $\num[round-mode = places]{78.38381499450907}$ & 
        $\num[round-mode = places]{31.142418111378483}$ &
        \bf{$\num[round-mode = places]{40.854223779457804}$} \\
        Softmax &
        $\num{15.812672681} \pm \numTwo{0.089978745}$  & 
        $\num{24.527707352} \pm \numTwo{0.053996400}$  & 
        $\num{78.315529844} \pm \numTwo{0.079793894}$  & 
        $\num{31.097286872} \pm \numTwo{0.028871124}$  & 
        $\num{40.509609850} \pm \numTwo{0.122710191}$  \\ 
        d=64, h=8 &  
        $\num[round-mode = places]{15.413995490859303}$ & 
        $\num[round-mode = places]{23.61514562159966}$ &
        $\num[round-mode = places]{78.48351849361724}$ & 
        $\num[round-mode = places]{30.96949603227393}$ &
        $\num[round-mode = places]{40.805487519474156}$ \\
        Softmax &
        $\num{15.897469640} \pm \numTwo{0.060322919}$  & 
        $\num{24.392476297} \pm \numTwo{0.028498873}$  & 
        $\num{78.381206911} \pm \numTwo{0.038577508}$  & 
        $\num{31.106734601} \pm \numTwo{0.017817047}$  & 
        $\num{40.662627653} \pm \numTwo{0.056110335}$  \\ 
        d = 128, h = 8 &
        $\num{15.940853825830132}$ &
        $\num{24.58933977382214}$ &
        $\num{78.29135730750848}$ &
        $\num{31.191942261508515}$ &
        $\num{40.62650993546146}$ \\
        Softmax &
        $\num{16.035070659} \pm \numTwo{0.080839093}$  & 
        $\num{24.714711934} \pm \numTwo{0.063554276}$  & 
        $\num{78.362449621} \pm \numTwo{0.074696344}$  & 
        $\num{31.196202485} \pm \numTwo{0.067497757}$  & 
        $\num{40.704445016} \pm \numTwo{0.053693214}$  \\ 
        \midrule
        \bf QANet+ELMo (Token) & \multicolumn{5}{l}{}    \\
        d = 64, h = 4 &  
        $\num[round-mode = places]{15.230794657823363}$ & 
        $\num[round-mode = places]{23.48239098958916}$ &
        $\num[round-mode = places]{78.246333058784}$ & 
        $\num[round-mode = places]{30.769323099246908}$ &
        $\num[round-mode = places]{40.2170837456049}$ \\
        Softmax &
        $\num{15.774599918} \pm \numTwo{0.145520634}$  & 
        $\num{24.441624727} \pm \numTwo{0.092950497}$  & 
        $\num{78.272501118} \pm \numTwo{0.032568643}$  & 
        $\num{31.062447538} \pm \numTwo{0.048212825}$  & 
        $\num{40.459904884} \pm \numTwo{0.105238931}$  \\ 
        d = 64, h = 8 &  
        $\num[round-mode = places]{15.302739194160967}$ & 
        $\num[round-mode = places]{23.407297463506286}$ &
        \bf{$\num[round-mode = places]{78.54211761134208}$} & 
        $\num[round-mode = places]{30.96686825341833}$ &
        $\num[round-mode = places]{40.18946979346518}$ \\
        Softmax &
        $\num{15.858178111} \pm \numTwo{0.068446490}$  & 
        $\num{24.397360279} \pm \numTwo{0.064433766}$  & 
        $\num{78.360976244} \pm \numTwo{0.075414933}$  & 
        $\num{31.111994054} \pm \numTwo{0.037515481}$  & 
        $\num{40.491637800} \pm \numTwo{0.050358916}$  \\ 
        d = 128, h = 8 &  
        $\num[round-mode = places]{15.238346336693425}$ & 
        $\num[round-mode = places]{23.58694013330103}$ &
        $\num[round-mode = places]{78.34104737988861}$ & 
        $\num[round-mode = places]{30.89584768210525}$ &
        $\num[round-mode = places]{40.18781952498374}$ \\
        Softmax &
        $\num{15.890561845} \pm \numTwo{0.075167358}$  & 
        $\num{24.552827022} \pm \numTwo{0.100807807}$  & 
        $\num{78.328864540} \pm \numTwo{0.058647350}$  & 
        $\num{31.110185304} \pm \numTwo{0.047704418}$  & 
        $\num{40.400958397} \pm \numTwo{0.053828640}$  \\ 
        \midrule
        \bf QANet+ELMo (Sentence) & \multicolumn{5}{l}{}    \\
        d = 64, h = 8 &  
        $\num[round-mode = places]{15.47603359579437}$ & 
        $\num[round-mode = places]{23.88134851854029}$ &
        $\num[round-mode = places]{78.43555387691627}$ &
        $\num[round-mode = places]{30.961283999884152}$ & 
        $\num[round-mode = places]{40.332224328956094}$ \\
        Softmax &
        $\num{15.995253684} \pm \numTwo{0.142636417}$  & 
        $\num{24.499412456} \pm \numTwo{0.330357113}$  & 
        $\num{78.341333969} \pm \numTwo{0.097839071}$  & 
        $\num{31.129990155} \pm \numTwo{0.083475562}$  & 
        $\num{40.562566876} \pm \numTwo{0.087607130}$  \\ 
        d = 128, h = 8 &  
        $\num[round-mode = places]{15.643673497249111}$ & 
        $\num[round-mode = places]{24.13239642151117}$ &
        $\num[round-mode = places]{78.51646731105834}$ &
        $\num[round-mode = places]{31.139380135359822}$ & 
        $\num[round-mode = places]{40.63224518131114}$ \\
        Softmax &
        \bf{$\num{16.054222562} \pm \numTwo{0.056937106}$}  & 
        \bf{$\num{24.807205017} \pm \numTwo{0.084232102}$}  & 
        $\num{78.398183357} \pm \numTwo{0.070499616}$  & 
        \bf{$\num{31.201663932} \pm \numTwo{0.056794468}$}  & 
        $\num{40.579787762} \pm \numTwo{0.029219118}$  \\ 
        \bottomrule
    
    \end{tabular}

    \label{tab:modelrescomb}
\end{table}

We can see in Table~\ref{tab:modelrescomb} that \emph{QANet with sentence-level ELMo} ($d_{model}=128$, $\#Heads=8$) performs best in terms of BLEU@2, ROUGE\_L, and Greedy Matching. Note also the correlation between higher results on the auxiliary task (see Table~\ref{tab:modelacc}) and improvement in terms of word-overlap measures, where we find the largest difference between individual and re-ranked models ($+1.5$ points absolute over the baseline, and $+0.95$ over seq2seq in terms of BLEU@2). In terms of semantic similarity, we note the highest increase for Embedding Average ($+1.3$ over the baseline, and $+1.4$ over seq2seq), and a smaller one for Greedy Matching ($+1.0$ over the baseline, and $+0.4$ over seq2seq), and Vector Extrema ($+2.6$ over the baseline, and $+0.6$ over seq2seq). 

Overall, the re-ranked models are superior as evaluated on word-matching measures, which is supported by the improvement of BLEU@2 and Embedding Average. The smaller improvement for Greedy Matching and Vector Extrema can be explained by the training procedure for the re-ranking model, which is based on word comparison. However, these two measures focus on keyword similarity between the target and the proposed answers, and generative models are better at this. This is supported by comparing the combined model to IR-BM25, where we see sizable improvements of $+1.5$ and $+2.0$ in terms of Greedy Matching and Vector Extrema, respectively.

We can further see in Table~\ref{tab:modelrescomb} that using a stochastic approach to select the best answer yields additional improvements. This strategy accounts for the predicted goodness score for each candidate, thus, enriching the model in two ways. First, implicit voting is used, as duplicate answer candidates are not removed, resulting in higher selection probability of popular answers from different input modules. Second, albeit two answers may have a very different structure, they still can be similar in meaning, leading to very similar scores and promoting only the first one. This behavior can be mitigated by choosing the winner proportionally to its ranking, thus, also introducing diversity in the chatbot's language. This hypothesis is supported by the results in Table~\ref{tab:modelrescomb}: compare each model to the corresponding one with \emph{softmax} selection.

\section{Conclusions and Future Work}
\label{sec:conclus}

We have presented a novel framework for re-ranking answer candidates for conversational agents. In particular, we adopted techniques from the domain of machine reading comprehension \cite{Chen:2017:DrQA, seo2016bi-directional, yu2018qanet} to evaluate the quality of a question--answer pair. Our framework consists of two tasks: ({i})~an auxiliary one, aiming to fit a goodness classifier using QANet and negative sampling, and ({ii})~a main task that re-ranks answer candidates using the learned model. We further experimented with different model sizes and two types of embedding models: GloVe~\cite{pennington2014glove} and ELMo~\cite{Peters:2018:ELMo}. Our experiments showed improvements in answer quality in terms of word-overlap and semantics when re-ranking using the auxiliary model. Last but not least, we argued that choosing the top-ranked answer is not always the best option. Thus, we introduced probabilistic sampling that aims to diversify the agent's language and to up-vote the popular answers, while taking their ranking scores into consideration.

In future work, we plan to experiment with different exploration strategies such as Boltzmann exploration
~({softmax is a degraded version of Boltzmann exploration with $\tau = 1$}).
 $\epsilon$-Confident, Upper confidence bound, and other bandit methods \cite{sutton:rl} to widen the possible context for each answer over time. We see an interesting research direction in applying deep reinforcement learning ({i})~to improve the answer selection models when applied to unseen questions, and ({ii})~to account for user feedback and customer support task success.

\vspace{6pt}

\authorcontributions{Conceptualization, M.H., P.N. and I.K.; Investigation, M.H; Writing--original draft, M.H.; Writing--review \& editing, P.N. and I.K.}

\funding{This research is partially supported by Project UNITe BG05M2OP001-1.001-0004 funded by the OP ``Science and Education for Smart Growth'', co-funded by the EU through the ESI Funds.}

\conflictsofinterest{The authors declare no conflict of interest.}

\reftitle{References}






\end{document}